\documentclass{article}

\usepackage{amsmath}
\usepackage{amsfonts}
\usepackage[margin=2.5cm]{geometry}
\usepackage[utf8]{inputenc}
\usepackage[english]{babel}
\usepackage{graphicx}
\usepackage{float}
\usepackage{tikz}
\usepackage{tikz-3dplot}
\usepackage{pifont}
\usepackage{array}
\usepackage{caption}
 
\usepackage[square,numbers]{natbib}
\usepackage{multicol}
\usepackage{parskip}
\newcommand{\win}[1]{\includegraphics[width=2cm]{#1}}
\renewenvironment{figure}
  {\par\medskip\noindent\minipage{\linewidth}}
  {\endminipage\par\medskip}
\renewenvironment{table}
  {\par\medskip\noindent\minipage{\linewidth}}
  {\endminipage\par\medskip}
\DeclareMathOperator{\sign}{sign}

\title{Convolutional Hashing for Automated Scene Matching}
\author{Martin Loncaric, Bowei Liu, Ryan Weber \\ Castle Global, Inc. \\ martin@thehive.ai}
\date{}

\begin{document}

\maketitle

\begin{multicols}{2}
\begin{abstract}
We present a powerful new loss function and training scheme for learning binary hash functions.
In particular, we demonstrate our method by creating for the first time a neural network that outperforms state-of-the-art Haar wavelets and color layout descriptors at the task of automated scene matching.
By accurately relating distance on the manifold of network outputs to distance in Hamming space, we achieve a 100-fold reduction in nontrivial false positive rate and significantly higher true positive rate.
We expect our insights to provide large wins for hashing models applied to other information retrieval hashing tasks as well.
\end{abstract}

\section{Introduction}

Many information retrieval tasks rely on high dimensional searches, including K-nearest neighbors (KNN), approximate nearest neighbors (ANN), and exact $r$-neighbor lookup in Hamming space.
At scale, these searches are enabled by indexes on binary hashes, such as locality-sensitive hashing (LSH) and multi-indexing \cite{multi12}.
Recent research has flourished on these topics due to enormous growth in data volume and industry applications \cite{big_data16}.
We present a powerful new approach to a fundamental challenge in these tasks: learning a good binary hash function.

We demonstrate the effectiveness of our method by applying it to the task of automated scene matching (ASM) with a multi-index system.
We call our model convolutional hashing for automated scene matching (CHASM).
To the best of our knowledge, it is the first neural network to outperform state-of-the-art hash functions like Haar wavelets and color layout descriptors at ASM across the board.

\begin{figure}
\label{wins}
\begin{small}\begin{sc}\begin{center}
\begin{tabular}{>{\centering\arraybackslash}m{4.2cm}>{\centering\arraybackslash}m{0.9cm}>{\centering\arraybackslash}m{2.3cm}}
frame pair & error type & benchmarks with error \\ \hline
\vspace{0.1cm}\win{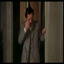} \win{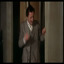} & FN & all \\
\win{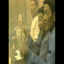} \win{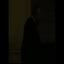} & FP & 192-bit CLD \\
\win{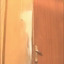} \win{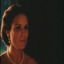} & FP & 64-bit wavelets, 256-bit wavelets \\ \hline
\end{tabular}
\end{center}\end{sc}\end{small}
\captionof{figure}{Select examples of false positives and false negatives that state-of-the-art hashes made but our 64-bit CHASM correctly handled.}
\end{figure}

\subsection{Automated Scene Matching}

ASM is an important information retrieval task, used to perform reverse video lookup for broadcasting, research, and copyright infringement monitoring \cite{cbvr15, german16}.
The goal of ASM is to take a query sequence of video frames and return all matching videos in a dataset, along with the start and end times of the matches\footnote{
ASM sometimes encompasses more than this specific definition. In particular, some ASM research aims to retrieve footage of the same 3D scene based on videos taken from another angle \cite{cbvir12}.
}.

For instance, suppose a research library indexes all their documentaries for reverse video lookup.
A researcher might then query the infamous Zapruder film, and her results should be all the documentaries containing a subset of it. The results should also include the specific time segments these documentaries matched the Zapruder film.

For large video datasets, this can be solved by implementing the following procedure\footnote{
Other procedures exist, especially ones that downsample to heuristic-selected keyframes rather than using a fixed frame rate.
However, these approaches are plagued by low recall and cannot distinguish between time granularities finer than their keyframes \cite{cbvr15}.
}:
\begin{enumerate}
\item For each video in the dataset:
\begin{enumerate}
\item Downsample to a fixed frame rate (fps) and image size.
\item Create a binary hash of each frame.
\item Using a multi-index lookup table, index each binary hash, pointing back to the source video and timestamp (Section \ref{multi_index}).
\end{enumerate}
\item For each query video:
\begin{enumerate}
\item Apply the same fps  and image size downsampling.
\item Create a binary hash of each frame.
\item Retrieve matches from the index for each binary hash.
\item Based on the individual frame matches, use heuristics to decide which dataset videos match during which time segments.
\end{enumerate}
\end{enumerate}

Our work optimizes the binary hash function used in steps 1b and 2b of this procedure. An ideal hash function for ASM must satisfy many requirements:
\begin{itemize}
\item Frames from the same video that are offset by up to a small time difference $t_0$ should map to hashes within the Hamming radius $r$ so that videos with a time shift still match together.
\item Frames offset by more time should map to hashes outside the Hamming radius so that the matching heuristics can determine precise start and end times.
\item Frames from different videos should map to hashes outside the Hamming radius to avoid false positive matches.
\item The false positive rate for each of the multi-index's indices must be extremely low, since the dataset may be very large, and each false positive increases query time and the probability of mismatching scenes.
\end{itemize}
It is worth noting that as dataset size increases, an ASM hash function's precision and recall drop, but its true positive and false positive rates stay the same.
Therefore we used true positive and false positive rates as our test metrics.

We compare our approach to state-of-the-art ASM methods, as well as variants of our own method, trained with other binary hash loss functions from recent research \cite{supervised16, hashnet17}.

\subsection{Multi-Indexing}
\label{multi_index}

Multi-indexing can enable search within a Hamming radius $r$ by splitting the $n$-bit hash into $r+1$ substrings of length $n/(r+1)$ \cite{multi12}\footnote{
In scenarios with a combination of extremely large datasets, short hashes, and large $r$, it may be more practical to use fewer than $r+1$ substrings and make up for the missing Hamming radius with brute-force searches around each substring \cite{multi12}. However, for ASM these conditions can be avoided by using larger hashes.
}.
Each of these substrings is inserted into its own index, pointing back to the full hash, video, and timestamp.

Lookup is performed as follows:
\begin{enumerate}
\item Taking an input hash $h$, split into substrings $h_1, \ldots h_{r+1}$.
\item Initialize an empty list $L$.
\item For $i=1, \ldots r+1$, add exact matches for $h_i$ in the $i$th index to $L$.
\item Filter duplicate results out of $L$.
\item Filter results with Hamming distance $>r$ out of $L$.
\item Return $L$
\end{enumerate}
The expected lookup runtime scales with $r + nm$, where $m$ is the number of exact matches per substring.
Therefore, with CHASM we seek to minimize not only the overall false positive rate, but also the false positive rate for each index.

\section{Related Work}

\subsection{Learning Binary Hash Functions}

Relevant to our method, some work has been done to find a good general method for learning binary hash codes.
Thus far these methods have relaxed discrete Hamming distance losses into differentiable optimizations by using piecewise-linear transformations on the hash embeddings \cite{minimal11, metric12}.
In this work we take these ideas further and leverage a more natural transformation.

\subsection{ASM}

So far neural networks have failed to outperform hand-picked features at hashing for ASM.
The main difference among existing state-of-the-art approaches comes from their hash functions, which are typically chosen from the frequency responses of some basis \cite{cbvr15}.
For wavelets, the discrete wavelet transform is run on images in grayscale, returning embeddings in the corresponding basis \cite{cbvr15}.
%For Gabor wavelets, the image is split into a 4x4 grid, then embedded in a Gabor basis \cite{cbvr15}.
The most common color descriptor representation is Color Layout Descriptor (CLD), which performs a discrete cosine transform on each channel of a smoothed image in YCbCr color space.
Each of these embeddings is generally binarized with a 1 for each above-median response and a 0 for the others \cite{cbvir12}.

%More recently, Schaffalitzky and Zisserman implemented an alternative approach for small datasets.
%Their technique involves computing 3D viewpoint invariant local features for each video keyframe, then comparing the features for each pair of keyframes \cite{asm02}.
%This approach proved very effective at matching keyframes from the same scene (i.e., the same background from a different angle).
%However, it is intractably slow for many industrial purposes; matching a single keyframe runs in $O(N)$ time (with a high constant), where $N$ is the number of keyframes in the dataset.

The state of ASM research leaves a major gap: learned methods that can perform temporally accurate scene matching quickly on very large datasets. In this paper, we used 3 benchmarks: the 64-bit (8x8) Haar wavelet hash, the 256-bit (16x16) Haar wavelet hash, and the 192-bit CLD hash.

\subsection{CBVR and CBIR}

Content based video retrieval (CBVR) is a broad topic that involves using video, audio, and/or metadata to retrieve similar videos from a dataset.
This is an easier task than ASM in that an entire video is retrieved, rather than a specific video segment.
There has been some recent research into learning a binary hash function for entire videos based on high-level, semantic labels \cite{video_deep17, video_mask17}.

Similarly, the objective of content based image retrieval (CBIR) is to take a query image as input and return a set of similar images in an image dataset.
Many recent papers in this field have also used deep learning approaches to train embeddings that get binarized into hashes.

So far, deep learning papers in these topics have mainly used a combination of three loss terms:
\begin{itemize}
\item terms that minimize or maximize the Euclidean distance between embeddings depending on whether they belong to similar or dissimilar content \cite{cnnh14, compact15, supervised16, efficient16, feature16, hashnet17, video_deep17, video_mask17, adversarial17}
\item classification loss terms that use a bottleneck before the classification layer as the hash layer \cite{fast15, semantic17, video_deep17, video_mask17}
\item binarization loss terms that punish embeddings for being far from $\pm1$ \cite{compact15, supervised16, efficient16, feature16, adversarial17}
\end{itemize}

Occasionally other loss terms are applied, including MSE from predefined target hash codes \cite{representation17} and adversarial error \cite{adversarial17}. 

We experimented heavily with these loss functions, but ultimately developed our own.
However, ideas from CBVR and CBIR papers such as using loss terms between each pair of images in a batch \cite{compact15} proved valuable in creating a good training scheme for CHASM.

Another notable trend in CBVR and CBIR research is the use of either binarization loss or learning by continuation \cite{hashnet17}; that is, gradually sharpening sigmoids to force embeddings close to $\pm1$.
This draws inspiration from the iterative quantization (ITQ) approach, which solves an alternating optimization problem of improving the embedding based on other metrics, then updating a rotation matrix to minimize binarization loss \cite{itq13}.
Unlike ITQ, more recent papers now allow gradients caused by binarization loss and learning by continuation to backpropagate through their network.

We find that backpropagating binarization loss or using learning by continuation causes learning to plateau, with the model only learning from a shrinking gray area of data points in between disconnected regions of data points near the corners of the hypercube $\{-1, 1\}^n$.
Moreover, the values in these regions do not binarize with the $\sign$ function any differently than less extreme values.
The main blocker preventing researchers from abandoning these methods is that Euclidean distance becomes a bad approximation for Hamming distance otherwise.

\section{Method}

\subsection{Interpretation of Embedding}

We propose an alternative to binarization loss and learning by continuation that respects the geometry of the embedding without punishing intermediate values.
We instead let our model produce embeddings following an approximately Gaussian distribution.

\subsubsection{Distribution of Embedding}

Let $x(f) = (x_1(f), \ldots x_n(f))$ be the vector of hash node outputs for an input frame $f$, and let $\mathcal{F}$ be the distribution of video frames to consider.
We motivate our loss function with the following assumptions:
\begin{itemize}
\item If $f \sim \mathcal{F}$ is a random video frame variable, then $x_i(f) \sim \mathcal{N}(0,1)$ (enforced by batch normalization of $x_i$ and a loss term on skew).
\item  $x_i$ is independent of other $x_j$.
\end{itemize}
Let $y(f) = x(f) / ||x(f)||_2$ be the $L_2$-normalized output vector.
Since $x(f)$ is a vector of $n$ independent random normal variables, $y(f)$ is a random variable distributed uniformly on the hypersphere.

This $L_2$-normalization is the same as SphereNorm \cite{sphere17} and very similar to Riemannian Batch Normalization \cite{riemannian17}.
Liu et al. posed the question of why this technique works better in conjunction with batch norm than either approach alone, and our work bridges that gap.
An $L_2$-normalized vector of IID random normal variables forms a uniform distribution on a hypersphere, whereas most other distributions would not.
An uneven distribution would limit the regions on the hypersphere where learning can happen and leave room for internal covariate shift toward different, unknown regions of the hypersphere.

\subsubsection{Estimate of Distribution of Hamming Distance}

To avoid the assumption that Euclidean distance translates to Hamming distance, we further study the distribution of Hamming distance given these $L_2$-normalized vectors.
We derive the exact probability that two bits match, given two uniformly random points $y^i, y^j$ on the hypersphere, conditioned on the angle $\theta$ between them.

\begin{figure*}
\label{distributions}
\begin{center}
\includegraphics[width=8cm]{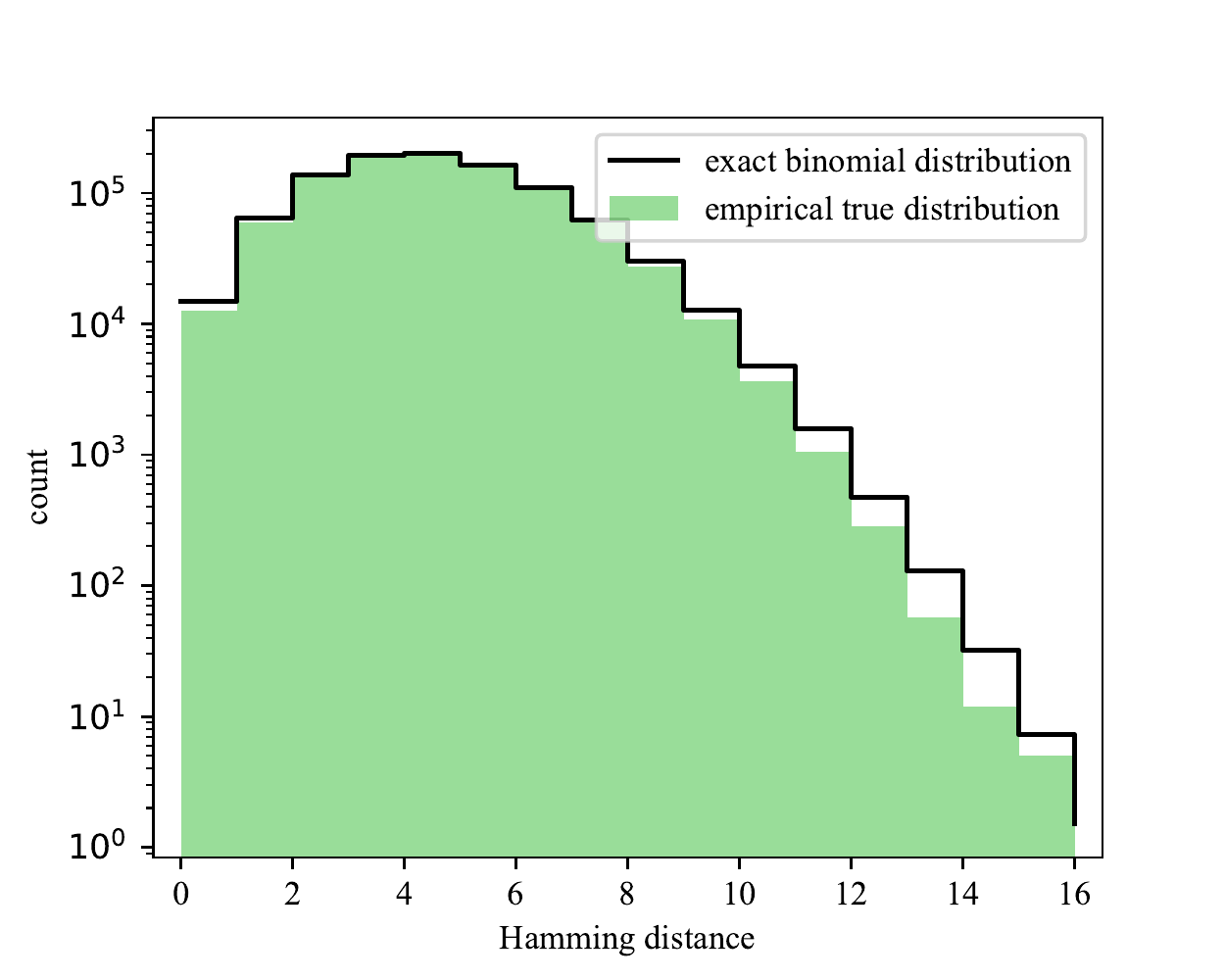}
\includegraphics[width=8cm]{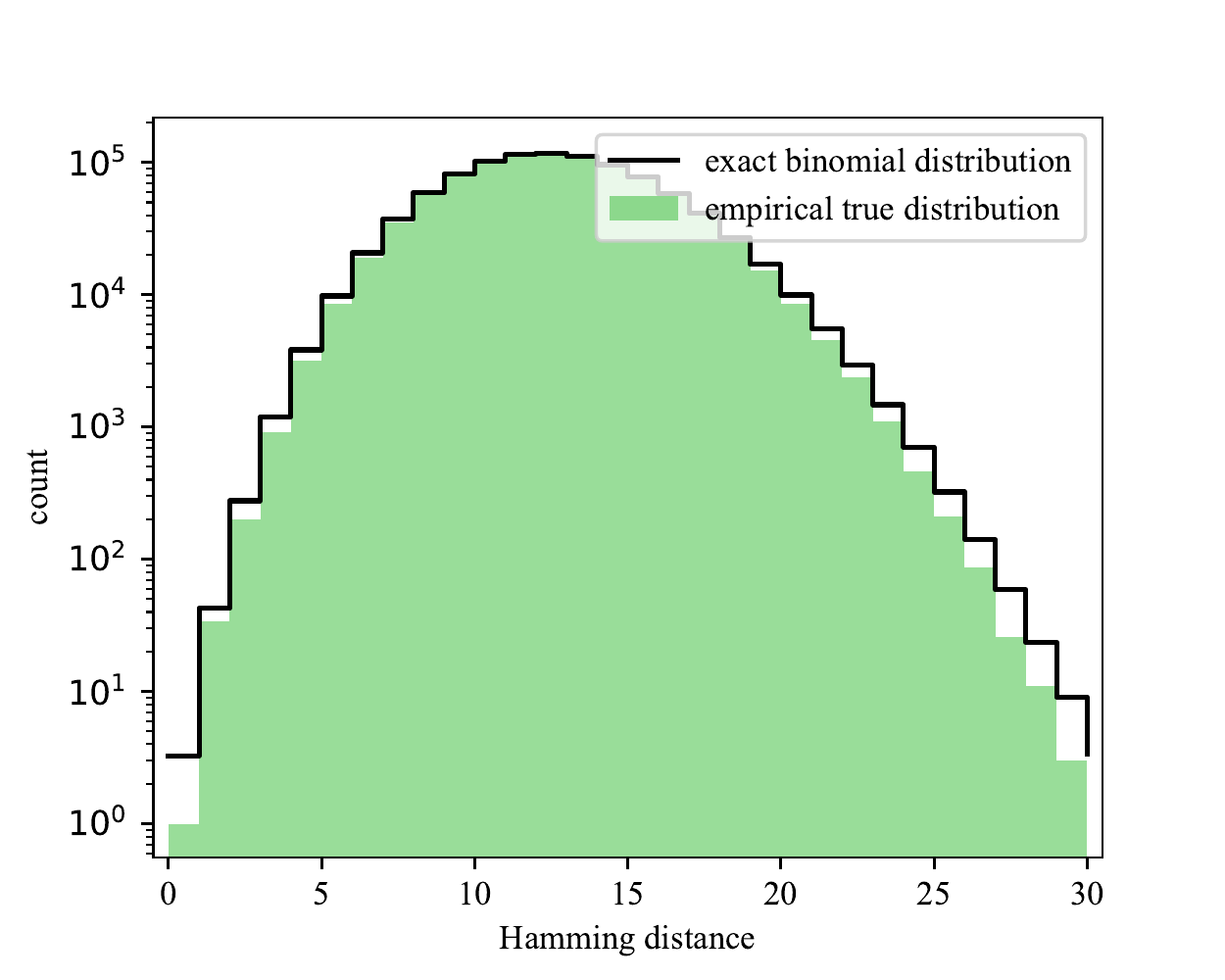}
\end{center}
\captionof{figure}{The distribution of Hamming distance for two uniformly random vectors on the $n$-hypersphere, conditioned on being separated by an angle $\theta = 0.2$. From left to right, $n=64,192$. Each empirical distribution was calculated from the results of $10^6$ trials.}
\end{figure*}

We know that $y^i \cdot y^j = \cos(\theta)$, so the arc length of the path on the unit hypersphere between them is $\arccos(y^i \cdot y^j)$.
A half loop around the unit hypersphere would cross each of the $n$ axis hyperplanes (i.e. $y_k = 0$) once, so a randomly positioned arc of length $\theta$ crosses $n \theta / \pi$ axis hyperplanes on average.
Each axis hyperplane crossed corresponds to a bit flipped, so the probability that a random bit differs between these vectors is
\[P^{ij} = \frac{\arccos\left(y^i \cdot y^j\right)}{\pi}\]
Given this exact probability, we estimate the distribution of Hamming distance by making the approximation that each bit position between the two vectors differs independently from the others with probability $P^{ij}$.
Therefore, the probability of Hamming distance being within $r$ is approximately $F(r; n, P^{ij})$ where $F$ is the binomial cumulative distribution function.
This approximation proves to be very close for large $n$ (Figure \ref{distributions}).

\tdplotsetmaincoords{20}{0}
\begin{figure}
\begin{center}
\begin{tikzpicture}[scale=3,tdplot_main_coords]
\draw [line width=1,domain=180:360, smooth] plot ({cos(\x)}, 0, {sin(\x)});
\draw [line width=1,domain=0:180,smooth,dashed] plot ({cos(\x)}, 0, {sin(\x)});
\draw [densely dotted] (0.5,-0.5,-0.707) -- (0,0,0);
\draw [densely dotted] (0.5,0.5,-0.707) -- (-0.5,-0.5,0.707);
%\draw [dashed] (0,0,0) -- (0.577,0,-0.816);
\draw [line width=0.5,domain=-75:30,smooth] plot ({0.577*cos(\x)}, {sin(\x)}, {-0.816*cos(\x)});
\draw [line width=0.5,domain=-150:-75,smooth,dashed] plot ({0.577*cos(\x)}, {sin(\x)}, {-0.816*cos(\x)});
\draw [line width=0.3,domain=-30:30,smooth] plot ({0.4*0.577*cos(\x)}, {0.4*sin(\x)}, {0.4*-0.816*cos(\x)}); 
\draw [line width=0.3,domain=-150:30,smooth] plot ({0.25*0.577*cos(\x)}, {0.25*sin(\x)}, {0.25*-0.816*cos(\x)}); 
\node at (0.3, 0, -0.4) {$\theta$};
\node at (-0.1, 0, -0.7) {$\pi$};
\node at (0.41, -0.49, -0.8) {$y^j$};
\node at (0.61, 0.47, -0.8) {$y^i$};
\node at (-0.61, -0.47, 0.8) {$-y^i$};
\draw [fill] (0.5, -0.5, -0.707) circle [radius=0.02];
\draw [fill] (0.5, 0.5, -0.707) circle [radius=0.02];
\draw [fill] (-0.5, -0.5, 0.707) circle [radius=0.02];

\tdplotsetrotatedcoords{90}{-20}{-90}
\draw [line width=2,domain=0:360,smooth,tdplot_rotated_coords] plot ({sin(\x)}, {cos(\x)}, 0);

\end{tikzpicture}
\end{center}
\captionof{figure}{
An arc of length $\theta$ on the unit hypersphere starting from a random point in a random direction has probability $\theta/\pi$ for the sign of a particular component to change along its course.
In the 3D example above, crossing the great circle implies that the sign of one component differs between $y^i$ and $y^j$.
}
\label{sphere}
\end{figure}
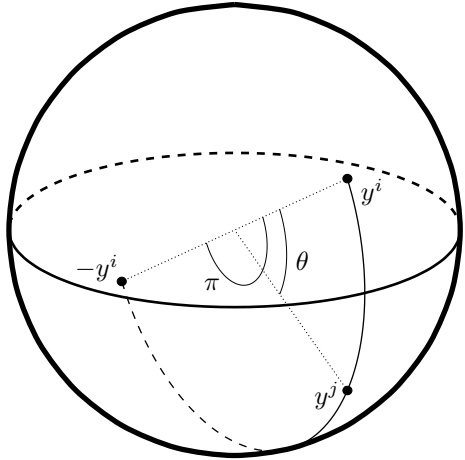

%\begin{figure}[H]
%\includegraphics[width=8cm]{Hamming_vs_dot.png}
%\caption{Our approximation for the Hamming correlation between two unbinarized hashes as a function of their Pearson correlation.}
%\label{Hamming_vs_dot}
%\end{figure}

Prior hashing research has made inroads with a similar observation, but applied it in the limited context of choosing vectors to project an embedding onto for binarization \cite{binomial16}. We apply this idea directly in network training.

\subsection{Classes of Time Differences}

For brevity, we define four classes of pairs of frames, depending on how far separated in time they are (Table \ref{hierarchy}). Our goal in CHASM is to maximize how often frame pairs in $H_0$ match together while minimizing how often frame pairs in $H_1, H_2,$ and $H_3$ do. Among $H_1, H_2,$ and $H_3$, it is by far most important that frame pairs in $H_3$ do not match together, since by far most frames in a video index will be from videos different than the query.

\begin{table}
\captionof{table}{Classes of frame pairs}
\begin{small}\begin{sc}\begin{center}
\begin{tabular}{cccc}
\hline
name & time difference & same shot & same video \\ \hline
$H_0$ & $\le t_0$ & yes & yes \\
$H_1$ & $>t_0$ & yes & yes \\
$H_2$ & $>t_0$ & no & yes \\
$H_3$ & $\infty$ & no & no \\ \hline
\end{tabular}\end{center}
\end{sc}
\end{small}
\label{hierarchy}
\end{table}

\subsection{Loss Function}

With batch size $b$, let $X = (x^1, \ldots x^b)^T$ be our batch-normalized logit layer for a batch of frames $(f^1, \ldots f^b)$ and $Y = (y^1, \ldots y^b)^T$ be the $L_2$-row-normalized version of $X$; that is, $y^i = x^i / ||x^i||_2$.
Similarly, let $Y_1, \ldots Y_{r +1}$ be the $b \times \frac{n}{r + 1}$ $L_2$-row-normalized submatrices formed by splitting $X$ into vertical slices; in other words, define a submatrix $Y_l$ for the logits of the $l$th substring.
Let $P = \frac{\arccos\left(Y^TY\right)}{\pi}$ and $P_l = \frac{\arccos\left(Y_l^TY_l\right)}{\pi}$.%, such that entry $P_l^{ij}$ corresponds to the probability that a random bit between frames $f^i$ and $f^j$ agrees.
Let $W$ be the vector of all our model's learnable weights.
Let $U_1, U_2, U_3$ be $b\times b$ matrices that depend on which class each pair of frames $(f^i, f^j$) is in (Table \ref{u_mats}).
We define our loss to be
\[J = -J_1 - J_2 - J_3 + \lambda_4J_4 + \lambda_5J_5\]
with
\begin{itemize}
\item $J_1 = \text{Avg}\left[U_1\circ\ln F\left(r; n, P\right)\right]$, the class-weighted average log likelihood of each pair of frames to be within Hamming distance $r$ (Table \ref{u_mats}).
\item $J_2 = \text{Avg}\left[U_2\circ\ln F\left(n - r - 1; n, 1 - P\right)\right]$, the class-weighted average log likelihood of each pair of frames to be outside Hamming distance $r$ (Table \ref{u_mats}). 
\item $J_3 = \sum_{j=1}^l\text{Avg}\left[U_3\circ\ln F\left(m - 1; m, 1 - P_j\right)\right]$, the class-weighted log likelihood that substrings differ between frames, summed over each substring (Table \ref{u_mats}). This term is particular to multi-indexing.
\item $J_4 = \frac{1}{nb}\left|\left|\sum_{i=1}^b(x^i)^3\right|\right|_2^2$, penalizing high skew and enforcing our assumption that the embedding follows a Gaussian distribution. We used $\lambda_4 = 2$.
\item $J_5 = ||W||_2^2$, a regularization term on the model's learnable weights to minimize overfitting. We used $\lambda_5 = 10^{-5}$.
\end{itemize}

%\begin{table*}
%\caption{Loss terms. $U_1, U_2, U_3$ are defined in Table \ref{u_mats}.}
%\begin{small}\begin{sc}\begin{center}
%\begin{tabular}{|c|c|c|}
%\hline
%term & formula & description \\ \hline
%$J_1$ & $\text{Avg}\left[U_1\circ\ln F\left(k; n, \frac{\arccos\left(YY^T\right)}{\pi}\right)\right]$ & log likelihood of within-radius frames \\
%$J_2$ & $\text{Avg}\left[U_2\circ\ln F\left(n - k - 1; n, 1 - \frac{\arccos\left(YY^T\right)}{\pi}\right)\right]$ & log likelihood of outside-radius frames \\
%$J_3$ & $\sum_{j=1}^l\text{Avg}\left[U_3\circ\ln F\left(m - 1; m, 1 - \frac{\arccos\left(Z_jZ_j^T\right)}{\pi}\right)\right]$ & log likelihood of partial indices \\
%$J_4$ & $\frac{1}{nb}\left|\left|\sum_{i=1}^b(x^i)^3\right|\right|_2^2$ & skew loss \\
%$J_5$ & $||W||_2^2$ & regularization term \\ \hline
%\end{tabular}
%\end{center}\end{sc}\end{small}
%\label{losses}
%\end{table*}

\begin{table}
\captionof{table}{Loss weights by frame pair class of $(f^i, f^j)$}
\begin{small}\begin{sc}\begin{center}
\begin{tabular}{ccccc}
\hline
Weight & $H_0$ & $H_1$ & $H_2$ & $H_3$  \\ \hline
$U_1^{ij}$ & 1 & 0 & 0 & 0 \\
$U_2^{ij}$ & 0 & 5 & $5\times10^2$ & $10^5$ \\
$U_3^{ij}$ & 0 & 0 & $10^2$ & $2\times10^4$ \\ \hline
\end{tabular}
\end{center}\end{sc}\end{small}
\label{u_mats}
\end{table}

Note that terms $J_1, J_2,$ and $J_3$ work on all pairwise combinations of images in the batch, providing us with a very accurate estimate of the true gradient.

\subsection{Dataset}

We trained our model using frames from Google's AVA video dataset \cite{ava17}, which consists of 154 training and 38 test videos annotated with activities.
For our purpose of automated scene matching, we disregarded the activity annotations.
We were able to obtain 136 of the training videos and 36 of the test videos.

To ensure that our model would learn meaningful similarities between frames, we selected the distinct cut-free shots of each video.
Then we filtered down to shots at least 4 seconds long and cut them to a maximum of 8 seconds.
We used a subset such that each was separated by at least 60 seconds from any other.
We then took training and testing shots from videos in the respective category, downsampling each shot at 15fps and $64\times64$ resolution to produce video frames.
We used all frames from the training set in training and distinct subsets from the test videos for validation and testing.

To find the cuts in each video, we used a cut detection model defined by \cite{Gygli17}.
We will make our shot annotations publicly available.
%future: link to github with csv of video segments used

\subsection{Architecture}

The network that learns the hash function is composed of 5 main blocks of convolutions (Table \ref{logonetplus}). 
Structurally it is similar to a wide resnet \cite{wideResnet} with the additional block added to handle the $64\times64$ input size.
Additionally, the pooling, classification, and softmax layers are removed and a fully-connected layer is added to specify the hash size.
By removing the global pooling, we allow the network to learn information about the position of features in images, which is important for automated scene matching.
We batch normalize the fully connected layer's outputs, giving the embedding.
From there, they are either $L_2$-normalized during training or binarized with the sign function during inference.

Following \cite{He2016IdentityMI} we used batch normalization before each convolutional layer and remove the activation function from the residual path.
In all our experiments, the depth factor was 6, which makes the network 49 convolutional layers.
We experimented with different width factors, but ultimately found no gains for width factors over 1.
This means the bulk of our resnet is identical to that of \cite{He2016DeepRL}.

\begin{table}
    \captionof{table}{Hash function architecture. Downsampling is performed by the first $3\times3$ convolution in each block with a stride of 2. 
    Batch normalization and ReLU activation precede each convolutional layer (except the first), and we add dropout between each
    convolutional weight layer.
We used $r=1$ and $N=6$.
}
\label{logonetplus}
\begin{small}\begin{sc}\begin{center}
\begin{tabular}{|c|c|c|}
  \hline			
  group name & output size & block \\ \hline
  conv1 & $64 \times 64$ & $\begin{bmatrix}3\times3,16 \times r \end{bmatrix}$ \\ \hline
  conv2 & $64 \times 64$ &  $\begin{bmatrix}3\times3,16\times r \\ 3\times 3,16\times r\end{bmatrix}\times N$ \\ \hline
  conv3 & $32 \times 32$ &  $\begin{bmatrix}3\times3,32\times r \\ 3\times 3,32\times r\end{bmatrix}\times N$ \\ \hline
  conv4 & $16 \times 16$ &  $\begin{bmatrix}3\times3,64\times r \\ 3\times 3,64\times r\end{bmatrix}\times N$ \\ \hline
  conv5 & $8 \times 8$ &  $\begin{bmatrix}3\times3,128\times r \\ 3\times 3,128\times r\end{bmatrix}\times N$ \\ \hline
      fc    & hash size & \\ \hline  
\end{tabular}
\end{center}\end{sc}\end{small}
\end{table}

\subsection{Training Scheme}

We chose $t_0$ to be $2/15$, such that 2 frames sampled at 15fps left or right of the query frame belong in $H_0$.

Using a batch size of $b=280$, we used what we call ``hierarchical batches'', which include pairs of images from each of $H_0, H_1, H_2,$ and $H_3$.
To construct one, we
\begin{itemize}
\item choose 35 random videos from our dataset with replacement,
\item choose 2 random shots from each video without replacement,
\item choose 3 random frames from each shot without replacement, and
\item choose 1 random additional frame within $r$ for each of those frames without replacement.
\end{itemize}
This ensures that even for very large datasets, each class is available enough to train on.

We trained our model using stochastic gradient descent with momentum for 8M images, or $s_0 = 28,600$  steps. Our network's weights randomly initialized to configurations with very high $J_2$ and $J_3$ loss terms, so we started our learning rate at a very low number $\beta$ for the first 1000 steps, gradually increasing until we began a cosine decay at a more typical learning rate $\alpha$:
\[\text{learning rate} = \begin{cases}
\beta \left(\frac{\alpha}{\beta}\right)^{s / 1000}, & s < 1000 \\
\alpha\left(1 + \cos\left(\frac{\pi s}{s_0}\right)\right), & s \ge 1000
\end{cases}\]
where $s$ is the global step.

To minimize overfitting, we used dropout with 30\% probability and flipped each batch of images horizontally with 50\% probability.

\section{Results}

\begin{table*}
\caption{Positive rate by frame pair class and hash function. $H_0$ and $H_3$ are by far the most important classes for these metrics. Values of $r$ were chosen by scanning the ROC curves of true positive rate vs. $H_3$ false positive rate for the best tradeoff (Figure \ref{roc}).}
\begin{small}\begin{sc}\begin{center}
\begin{tabular}{|c|c|c|c|c|c|}
\hline
Hash Function & chosen $r$ & $H_0$ TP rate & $H_1$ FP rate & $H_2$ FP rate & $H_3$ FP rate \\ \hline
Haar Wavelets (64-bit) & 3 & 0.810 & 0.261 & $1.48\times10^{-4}$ & $2.18\times10^{-5}$ \\
Haar Wavelets (256-bit) & 14 & 0.834 & 0.265 & $5.46\times10^{-5}$ & $1.75\times10^{-5}$ \\
CLD (192-bit) & 16 & 0.835 & 0.265 & $4.12\times10^{-5}$ & $1.76\times10^{-5}$ \\
CHASM (64-bit) & 3 & 0.885 & 0.334 & $3.16\times10^{-5}$ & $5.00\times10^{-6}$ \\
CHASM (192-bit) & 7 & $\mathbf{0.8877}$ & 0.319 & $\mathbf{2.04\times10^{-6}}$ & $\mathbf{4.83\times10^{-6}}$ \\
CHASM-C (192-bit) & 8 & 0.621 & $\mathbf{0.115}$ & $9.73\times10^{-5}$ & $6.65\times10^{-5}$ \\
CHASM-B (192-bit) & 1 & 0.800 & 0.282 & $2.17\times10^{-3}$ & $9.81\times10^{-4}$ \\
CHASM-N (192-bit) & 7 & 0.878 & 0.288 & $1.48\times10^{-5}$ & $5.67\times10^{-6}$ \\
%CHASM (192-bit) & \textbf{0.946}  & 0.448 & $8.18\times10^{-5}$ & $5.67\times10^{-6}$ \\
%CHASM (192-bit) & 32 & & & & \\
\hline
\end{tabular}
\end{center}\end{sc}\end{small}
\label{losses}
\end{table*}

\begin{figure}
\includegraphics[width=9cm]{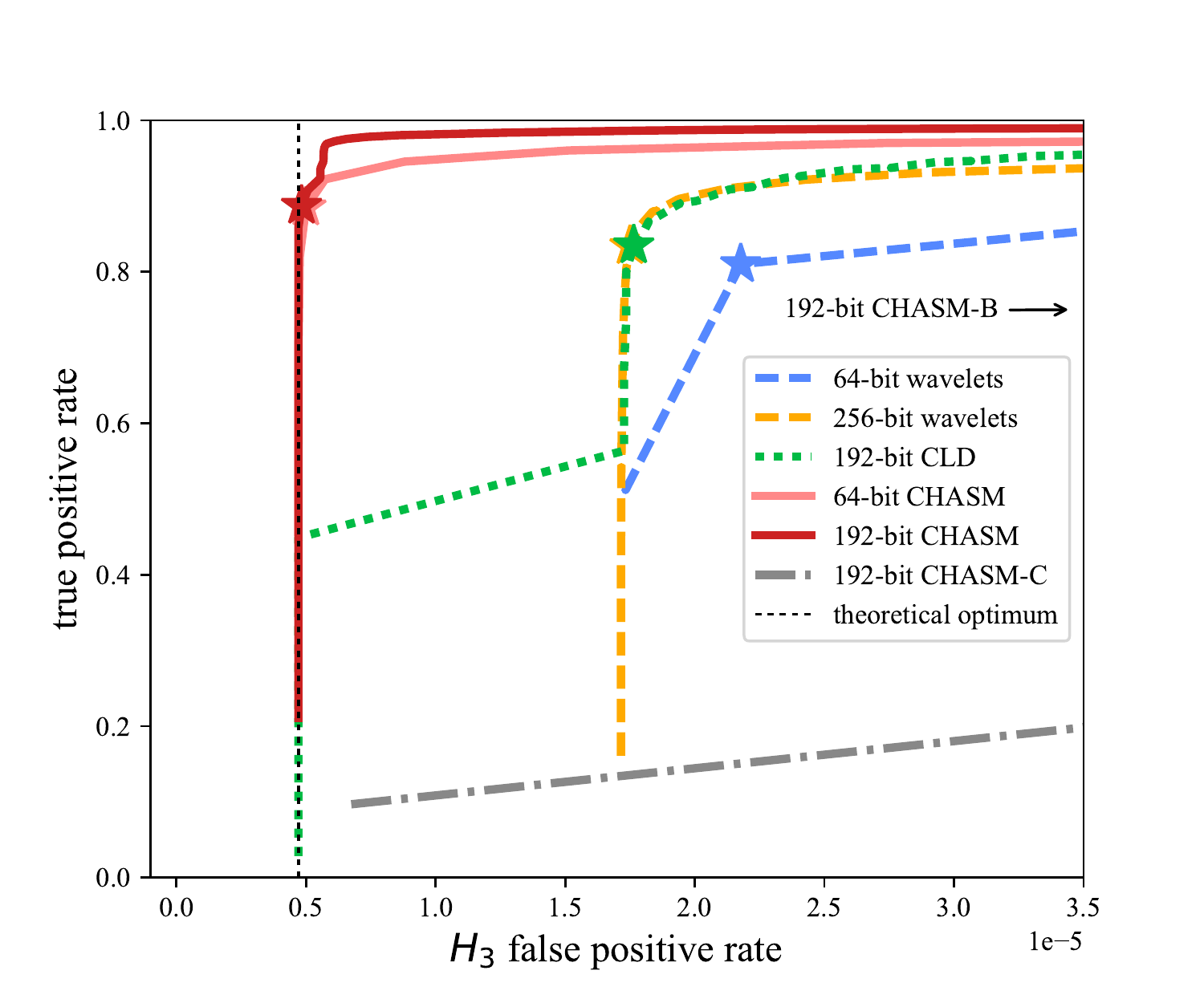}
\captionof{figure}{
True positive rate vs. false positive rate among frames from different videos, plotted at different values of $r$.
Each star corresponds to a heuristically chosen value of $r$ on this dataset that has at least 60\% TP rate and maximizes $\text{TP rate} - 10^5\text{FP rate}$.
CHASM-N is omitted due to clutter.
}
\label{roc}
\end{figure}

We trained CHASM models for hashes of 64 and 192 bits, optimizing for binary substrings size of 32.
We tested our results on the over $1.3\times10^9$ distinct pairs of frames in our test set.
CHASM achieved higher true positive rates and lower false positive rates for each class and hash size; in fact, even our 64-bit hash beat the 192- and 256-bit benchmark hashes on both true positive rate and $H_3$ false positive rate by a large margin (Figure \ref{roc}).

The lowest possible $H_3$ false positive rate on our test set was $4.71\times10^{-6}$, since various videos contained perfectly identical black frames.
These results can be avoided in practice by ignoring any perfectly black frames, which are not very informative.
Without these frames, the 192-bit CHASM achieves $1.2\times10^{-7}$ nontrivial $H_3$ false positive rate at 88.8\% true positive rate, over a 100-fold reduction of the $1.31\times10^{-5}$ nontrivial $H_3$ false positive rate of the best benchmark hash (256-bit wavelets) at 85.9\% true positive rate.

In addition, we compare against two variants of CHASM, modified by removing batch and $L_2$ normalization on the embedding and using different loss functions:
\begin{itemize}
\item CHASM-B, using a squared error loss term between each frame pair depending on similarity and an $L_1$ binarization loss term for how far the embedding is from binary.
\item CHASM-C, using a logistic loss term based on the dot product of embeddings. We also used learning by continuation, computing the embeddings by passing our final layer through a $\tanh$ layer that periodically gets sharper throughout training.
\end{itemize}
We implemented the loss function from \cite{supervised16} for the former and that of \cite{hashnet17} for the latter, along with appropriately tuned learning rate schedules and hyperparameters.

Neither approach performed on the same level as any of our benchmark hashes, let alone CHASM (Table \ref{losses}).
However, this is not in contradiction with their respective papers' results; both worked for low-recall, high-precision tasks like finding nearest neighbors.

Finally, we compared against a model CHASM-N trained without skew loss ($J_4$). We found that it had lower true positive rate and higher $H_3$ false positive rate for every value of $r \le 16$.

\section{Conclusion}

Our results show for the first time that a neural network is capable of outperforming traditional hashing methods at the task of hashing for ASM.
Our model was able to reduce nontrivial false-positive rate on a large dataset by a factor of 100, even at higher true positive rate.
This constitutes a massive improvement in the speed and accuracy of ASM systems.

In contrast, we found that state-of-the-art approaches to CBIR were unable to beat even our benchmarks.
We attribute our model's comparable success to four main factors.
\begin{itemize}
\item CHASM's loss depends on the chance of misclassifying an image.
\item CHASM uses good estimates for the distribution of Hamming distance as a function of embeddings.
\item CHASM does not restrict the embedding's values near $\pm1$ during training, which (without actually changing its binarized values) can prevent the model from learning.
\item CBIR research has focused on low-recall, moderate-precision regimes like nearest neighbors, whereas ASM demands extremely low false positive rate.
\end{itemize}

We also shed light on why $L_2$-normalization of layer outputs improves learning in conjunction with batch norm.
Perhaps most importantly, we provide a powerful new loss function and training scheme for learning binary hash functions in general.

\bibliography{ArxivChasm}
\end{multicols}
\end{document}